# MaCmS: Magahi Code-mixed Dataset for Sentiment Analysis

**Priya Rani**[1], **Gaurav Negi**[1], **Theodorus Fransen**[2], **John P McCrae**[1]

[1] Data Science Institute, University of Galway, Ireland
[2] Università Cattolica del Sacro Cuore, Milan, Italy
{priya.rani, gaurav.negi, John.mccrae}@insight-centre.org
theodorus.fransen@unicatt.it

**Abstract**

The present paper introduces new sentiment data, MaCmS, for Magahi-Hindi-English (MHE) code-mixed languages, where Magahi is a less-resourced minority language. This dataset is the first Magahi-Hindi-English code-mixed dataset for sentiment analysis tasks. Further, we provide a linguistic analysis of the dataset to understand the structure of code-mixing and a statistical study to understand the language preferences of speakers with different sentiment categories. With these analyses, we also train baseline models to evaluate the dataset's quality.

## 1. Introduction

Different people can perceive a word differently, and these differences are rarely distinctive; however, these are often linked with social factors, such as age, gender, race, geography and more inferable characteristics, such as political and cultural attitudes (Yang and Eisenstein, 2017). Sociolinguists and psychologists have been studying these variations in the lexicons and the language from the 50's (Fischer, 1958; Labov, 1963). The advent of social media has created extensive opportunities to explore these variations automatically. In multilingual societies, particularly on social media, code-mixing, which involves using multiple languages within a single conversation or text, is a well-known phenomenon. This practice has expanded due to the global reach of social media, making it a valuable source of texts for less-resourced languages. One of the applications of this data is sentiment analysis, which seeks to understand people's emotions and attitudes towards various subjects. However, this becomes a challenge for low-resourced languages, primarily because text data in these languages often includes a mix of different scripts and languages.

Sentiment analysis is commonly regarded as a task involving categorising text into one of three categories: positive, negative, or neutral, as mentioned in Phani et al. (2016) study. In recent years, due to the proliferation of social media platforms like YouTube and Twitter, sentiment analysis has expanded into a broader range of applications in various fields. Multinational companies can use it to analyse customer feedback and reviews. In politics, it can gauge public sentiment towards candidates and policies. For social media platforms, it can be used to monitor and filter out offensive content. Sentiment analysis not only reveals the mood and emotional state of the speaker but also provides insights into cultural and political attitudes. This information is invaluable for organisations, governments, and businesses in understanding public perception and sentiment towards their products, policies, or initiatives. The brevity and informality of social media posts, like tweets and YouTube comments, present unique challenges and opportunities for sentiment analysis. Short text analysis can be more challenging than longer texts, as less context might be available for determining sentiment. Moreover, issues like code-switching and the unavailability of resources for less-resourced languages can complicate the tasks further (Kenyon-Dean et al., 2018).

While sentiment analysis as a field has been expanding, and numerous systems have made remarkable progress in setting new performance standards, the effectiveness of sentiment prediction in the context of code-mixed data still needs to be improved. This limitation is primarily attributed to the variability in language availability and the quality of training data, which directly impacts the precision of sentiment analysis. Sentiment analysis for Indian languages, especially in code-mixed settings, is still relatively nascent (Jose et al., 2020; Chakravarthi et al., 2020a,b; Rani et al., 2020). The significant difference in style of language, orthography (Chakravarthi et al., 2019) and grammar used in tweets presents specific challenges for the Indian languages code-mixed data especially when it comes to low-resource languages like Magahi. Therefore, this work presents a sentiment analysis dataset for Magahi-Hindi-English code-mixed language. Our major contributions to the study are as follows:

- MaCmS: an annotated Magahi (MAG), Hindi (HIN), English (ENG) code-mixed dataset for sentiment analysis. To the best of our knowledge, this is the first Magahi code-mixed dataset for sentiment analysis.

- A linguistic analysis of the structure of code-mixing between two closely related languages, Magahi and Hindi. Which helps understand when and where the code-mixing is happening.

- A statistical analysis of the dataset based on the language preference used by the speakers, which indicates the emotions and attitude of the speakers and the sentiment they show.

- We also provide some baseline models for sentiment analysis at the sentence and language-specific span levels.

## 2. Literature Review

There is abundant research work on sentiment analysis. These works study the diverse information about individuals on social media platforms. The primary goal of these studies is to gain insights into the emotions experienced by individuals, which play a fundamental role in personal development, intellectual growth, and human understanding (Xia et al., 2020). Even though the past decade has taken the study of sentiment analysis to a more advanced level, it still lacks in the case of less-resourced languages, and the performance of sate-of-the-art is yet to be surpassed in different multilingual settings (Ali et al., 2021).

Several resources have been created for sentiment analysis over the period. Yu et al. (2020) introduced a Chinese single- and multi-modal sentiment analysis dataset, CH-SIMS, which contains 2,281 refined video segments in the wild with both multimodal and independent unimodal annotations. XED, a multilingual fine-grained emotion dataset which consists of human-annotated 25 K Finnish and 30 K English sentences, as well as projected annotations for 30 additional languages, providing new resources for many low-resource languages (Öhman et al., 2020). Patwa et al. (2020) organised a shared task on sentiment analysis, leading to the release of the Hindi-English and Spanish-English language pairs dataset.

With the development of practical sentiment analysis guidelines by Mohammad (2016) led to the development of many corpora. Even the vast language diversity in the Indian subcontinent led to the creation of sentiment corpora for many of the less-resourced languages. Ali et al. (2021) created a Sindhi subjective lexicon with the help of existing English resources. This resource consists of opinion words, SentiWordNet, a Sindhi-English bilingual dictionary and collection of Sindhi Modifiers. Each opinion word is paired with a positive or negative sentiment score. Chakravarthi et al. (2020b) developed a Tamil-English sentiment analysis corpus with 15,744 comment posts from YouTube. The authors also provide the baseline models for the gold standard dataset. Ehsan et al. (2023) developed BiLSTM-based models to train sentiment analysis classifiers for Tamil and Tulu code-switched datasets by deploying contextualized word embeddings at input layers of the models. The F1-score of the models are 0.2877 and 0.5133 for Tamil and Tulu datasets, respectively.

Patra et al. (2015) organised the shared task competition on sentiment analysis of code-mixed data pairs for Hindi-English and Bengali-English. The best-performing team used SVM for sentence classification. Kulkarni et al. (2021) published L3CubeMahaSent, the first publicly available Marathi sentiment analysis dataset. It consists of 16,000 distinct tweets classified into three classes: positive, negative, and neutral, with extensive baseline models using CNN, LSTM, ULM-FiT, and BERT. Gokani and Mamidi (2023) describes the Gujarati Sentiment Analysis Corpus (GSAC), sourced from Twitter and manually annotated by native speakers. The paper discussed the creation process of the dataset and provided extensive baseline experiments with the highest F1-score of 0.66 from the IndicBert model.

## 3. MaCmS Dataset

### 3.1. Data Creation and preprocessing

YouTube is a popular and readily available source for collecting Magahi corpus. We gathered the data from publicly available comments on YouTube as some Magahi speakers are highly active on YouTube. For this task, we selected two YouTube channels, namely, 'Magadhi Boys'[1] and 'Magadh Music' [2]. As the study conducted by Rani et al. (2022) these two channels are very active, and the videos uploaded are on various themes such as folklore, mythology, society, politics, environment, and entertainment, providing us leverage to study the attitudes and emotions of the Magahi speakers. Moreover, being a minority and a less-resourced language, Magahi has a minimal presence on social media, thus giving us less opportunity to collect digital data. Further, we removed the duplicate comments from the data and labelled the polarity of the comments at the sentence level. Secondly, We tokenised the comments into language-specific spans and tagged these spans with their polarity.

---

[1] https://www.youtube.com/channel/UCvh5PbwK8I3lyRSQSjsqYwQ
[2] https://www.youtube.com/c/MagadhMusic1

## 3.2. Data Annotation

The dataset aimed to get the polarity of the comments for sentiment analysis in closely related code-mixed text for low-resourced settings. At the same time, we are also using the dataset to get language-specific features from the language-specific span and its polarity. This dataset includes annotations from four annotators. Two of the four annotators were female, and the other two were male. Among the two female annotators, one is a trained linguist, and the other is a computer science student. Among the two male annotators, one is a language student, and the other is a mathematics student. However, all four annotators' mother tongue was Magahi, and they were fluent in all three languages involved in the dataset. They are aged 19-28 and belong to middle to upper-middle-class families. The four annotators originally belong to Bihar, India, where Magahi is one of the dominant languages. We asked the annotators to annotate the polarity of the text at two levels: sentence level and language-specific span discussed below:

- Sentence Level: We label a comment's polarity at the sentence level.

- Span level: Where each comment is tokenised with language-specific span or, in simple terms, we tokenise the code-mixed text into language-specific tokens from its code-mixing points. The tokenisation is done manually with the help of native speakers.

| Number | Sentence | Translation | Sentiment |
|---|---|---|---|
| Sent: 1 | Bahut sundar bahut wait kara hi apne vedio ke | Very nice I am eagerly waiting for your video | Positive |
| Sent: 2 | तोरा दोनो के सामने सलमान खान शाहरुखखान आमीर खान अजय देवगनसब फैल है जी हमको बिहारी होने पे इतना गरब है कि बता नय सझीवो हा जी तोर दोनो के वीडियो देखो हीवो तो मन खुश हो जाहो | Salman Khan, Shahrukh Khan and Ajay Devgan are nothing in front of you both; I am so proud to be a Bihari that I can't express. Moreover, I am delighted whenever I see your videos | Positive |
| Number | Span and (language) | Translation | Sentiment |
| Sent: 1 Span1 | Bahut sundar bahut (Hindi) | Very nice | Positive |
| Span2 | wait (English) | wait | Neutral |
| Span3 | kara hi apne vedio ke (Magahi) | For your video | Neutral |
| Sent: 2 Span1 | तोरा दोनो के सामने सलमान खान शाहरुखखान आमीर खान अजय देवगनसब फैल है जी (Magahi) | Salman Khan, Shahrukh Khan and Ajay Devgan are nothing in front of you both | Positve |
| Span2 | हमको बिहारी होने पे इतना गरब है कि (Hindi) | I am so proud to be bihari that | Positive |
| Span3 | बता नय सझीवो हा जी तोर दोनो के वीडियो देखो हीवो तो मन खुश हो जाहो (Magahi) | I can't express. Moreover, I am delighted whenever I see your videos. | Positive |

Figure 1: Demonstrations of the annotation with examples.

## 3.3. Annotation Guidelines

**Positive Sentiment (POS):** A sentence or a language-specific span of a comment is labelled with positive polarity with the tag `POS` when the text reflects the following criteria:

- The speaker uses positive language to express support, admiration, positive attitude, forgiveness, fostering, success, and a positive emotional state.

- The choice of annotation should remain independent of agreeing or disagreeing with the speaker's opinions. Evaluators should focus on assessing the language rather than judging the views expressed.

- The text contains directly or indirectly implied indications that the speaker is experiencing positive emotions such as happiness, admiration, relaxation, and forgiveness.

**NEG: Negative Sentiment**

- The speaker uses negative language, for example, expressions of criticism, judgement, negative attitude, questioning validity/competence, failure, and negative emotion.

- There is an explicit or implicit clue in the text suggesting that the speaker is in a negative state.

- The speaker is using expressions of sarcasm, ridicule or mockery.

**NEU: Neutral Sentiment**

- The speaker is using neither positive nor negative language

- There is no explicit or implicit clue of the speaker's emotional state indicating the speaker feels positively or negatively

**MIX: Mixed sentiment**

- The speaker is using positive and negative language in part.

- There are explicit and implicit clues in the text suggesting that the speaker has positive and negative feelings.

## 3.4. Workflow of annotation

The annotation was accomplished in three phases. In the first phase, we gave each annotator a sample of 500 sentences and 100 spans. After the annotation, we manually evaluated the annotation.

We made some subtle changes in the annotation guidelines with a more detailed explanation of the tag and providing examples as references with each annotation guideline. The initial guidelines were motivated by the practical guide to sentiment analysis discussed by Mohammad (2016). After the first annotation phase, we had a feedback session with the annotators to improve the annotation quality and clearly understand the annotation guidelines. The second phase of the annotation was again conducted with another set of samples with 500 sentences and 100 spans, where we found much improvement in the annotation task. At last, when we were satisfied with the annotators, we gave them the data for the final round of annotation with a set of 1000 sentences and 500 spans one at a time. With each batch of annotation, we manually evaluated the sentiment of each annotator, and in disagreement, we followed the rules mentioned below to get the final decision.

- When there is a disagreement between all the annotators, we re-annotated the text by ourselves.

- If there is disagreement by one annotator but the other three annotators agree on a tag, then the final tag of the text would be the tag selected by the majority. For example, if three of the annotators tagged a text or a span with `POS` and one of the annotators tagged it as `NEU`, then the final tag would be `POS`

- If two annotators agree and the other two have two different opinions about the polarity of the text, then the final tag was given with majority voting. For example, if a text is labelled with `POS` by two annotators while the third annotator labelled it as `NEG` and the fourth annotator marked it as `NEU`, then in this case, the final label would be `POS`.

- When the agreement between the annotators is distributed equally, for example, if two annotators labelled the text with `POS` and the other two annotators labelled it as `NEU`, we sent the text for re-annotation.

## 4. Data Analysis

### 4.1. Inter-annotator Agreement

After each annotation phase, we calculated the inter-annotator agreement to check the annotation process's reliability and ensure the dataset quality. We used Krippendorff's $\alpha$ to estimate the agreement between the annotators. The agreement score helped us visualize the improvement we need to make in the annotation guidelines to improve the quality of the annotated dataset. Table 1 shows the obtained IAA between the annotators at each phase. Considering the nature of the task is very subjective, we achieved a fair final agreement score between the annotators, that is, 0.78 for sentence-level annotation and 0.76 for span-level annotation.

| Phase | Sentence-level | Span-level |
|---|---|---|
| Phase-1 | 0.67 | 0.69 |
| Phase-2 | 0.72 | 0.76 |
| Final | 0.78 | 0.76 |

Table 1: Krippendorff's $\alpha$ for inter-annotator agreement of the MHE sentiment analysis dataset

### 4.2. Data Statistics

We collected 11000 comments from the mentioned YouTube Channels. However, we discarded the comments in English or Hindi, and at last, we were left with 5663 comments. Out of these, 5,000 comments were labelled for sentence sentiment analysis. We tokenised the comments for language-specific span. Due to the annotators' time constraints, we could annotate only 750 sentences for span-level sentiment analysis, which resulted in 2642 spans. Table 2 gives brief statistics of the dataset.

| MHE | Number |
|---|---|
| Sentences | 5000 |
| Span sentences | 750 |
| Total span | 2642 |

Table 2: Statistics of our YouTube Magahi-Hindi-English (MHE) code-mixed dataset for sentiment analysis

The distribution of the sentiment tags for sentence-level analysis is demonstrated in Figure 2. We find that the distribution of positive and negative tags is mostly balanced, with 37.4% and 33.8%, respectively. However, the neutral and mixed sentiment percentage is low, with 16.2% and 12.6%, respectively. In contrast, the rate of neutral tags in the language-specific span exceeds and consists of 50.2% of the whole dataset, as shown in Figure 3. We also find that the span with positive tags acquires the second position with 35.0% of the entire sentiment dataset, followed by negative tags with 9.8% and mixed tags with 5%.

## 5. Baselines Experiments

This section will briefly describe the various models used for the baseline experiments. We conduct a very limited pre-processing that removes any URLs from the data before or after dividing the data into the training, validation, and test sets with the distribution of 70%, 20% and 10%, respectively. However, no further pre-processing steps, such as casing or data normalisation, are conducted to make the baseline experiments more robust.

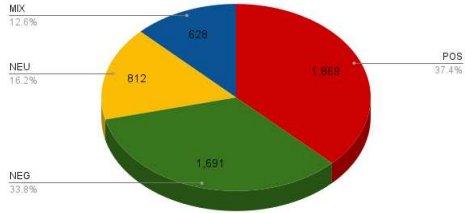

Figure 2: Distribution of the tags in the MHE dataset for sentence-level sentiment analysis.

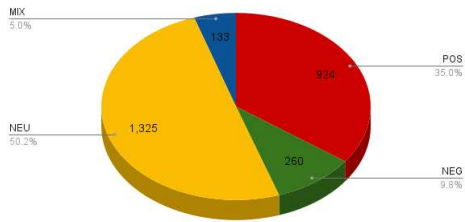

Figure 3: Distribution of the tags in the MHE dataset for language-specific span-level sentiment analysis.

### 5.1. mBERT

We have used the multilingual pre-trained BERT (Devlin et al., 2019; Turc et al., 2019) model to fine-tune our first baseline model. It has an embedding dimension of 768. We have trained our model for 10 epochs on the training data set and have used a stepped LR scheduler for the learning rate schedule. The learning rate is set to 2e-5. Based on the Hugging Face implementation, we have used the below equation 1 as a warm-up step definition for training the model. Here, 'r' is the tuneable parameter, which defines the percentage of data used to define the step size while training.

$$W_{steps} = \frac{(len(training_{set}) \times epochs_{training})}{batchsize_{training} \times r}  \quad (1)$$

### 5.2. XLM-R

We use XLM-RoBERTa (XLM-R) (Conneau et al., 2020) to fine-tune the second model. As XLM-R is pre-trained on 100 languages, which includes Indian languages such as Hindi and Hindi being one of the closely related languages to Magahi, it would provide some aid to train the model on the Magahi-Hindi-English dataset for sentiment analysis. As shown in the previous research (Winata et al., 2021) that the XLM-R model performs better than other multilingual models, we investigate the effectiveness of XLM-R in less-resource language settings. We used pre-trained models from Hugging Face[3]. We put a fully connected classifier on each model and trained the model with a learning rate of 1e-5 with a decay of 0.1 and batch size of 32 for 10 epochs.

### 5.3. GenMA

Generative Morphemes with Attention (GenMA) model (Goswami et al., 2020) is a sentiment analysis model trained to classify the sentiment based on the newly generated artificial character sequences termed as artificial morphemes. This model combines two convolution layers with one max pooling layer, a BiLSTM layer and an attention layer. As the model generates new character sequences irrespective of language and orthographic features, it helps to capture both language-specific and code-mixed language features. Thus, this model is used to train sentiment classifiers on less-resource languages.

We have used 32 filters, each with a kernel size of 3. The max-pooling size is 3. The hidden size of LSTM units is kept to 100. The dense layer has 32 neurons and a 50 percent dropout. The Adam optimizer (Kingma and Ba, 2015) trains the model with the default learning set to 0.0001. The batch size is set to 10. We have used the relu activation function (Nair and Hinton, 2010) and tanh activation function (Kalman and Kwasny, 1992) for the convolution and dense layer, respectively. Categorical cross-entropy loss is used for the multi-class classification.

## 6. Results

Overall, we see varying performance across the models, with some performing much better than others. We computed the F1 score, Precision and Recall to evaluate the performance of the models on our dataset to account for the imbalance in label distribution (see Figure 3). The evaluation scores of the models are summarised in Table 3 for sentence-level sentiment analysis and Table 4

---
[3] https://huggingface.co/

for language-specific span-level sentiment analysis.

| Models | Precision | Recall | F1 Score |
|---|---|---|---|
| mBERT | 0.65 | 0.71 | 0.68 |
| XLM-R | 0.76 | 0.75 | **0.75** |
| GenMA | 0.68 | 0.69 | 0.68 |

Table 3: The evaluation results of the models at sentence level sentiment analysis. The bold results reflect the best performance of the model.

| Models | Precision | Recall | F1 Score |
|---|---|---|---|
| mBERT | 0.55 | 0.52 | **0.53** |
| XLM-R | 0.50 | 0.48 | 0.49 |
| GenMA | 0.51 | 0.50 | 0.51 |

Table 4: The evaluation results of the models at span level sentiment analysis. The bold results reflect the best performance of the model.

Looking at the evaluation table, we can observe that XLM-R performed the best for sentence-level sentiment analysis with the F1-score of 0.75, while mBERT F1-score was the highest for span-level sentiment analysis. The poor performance of the span-level models is due to the lack of training data. These results also evaluate the quality of the dataset. As we know, these large language models are data-hungry. They performed comparatively well in code-mixed and less-resourced settings, concluding that the data annotation quality is good.

## 7. Discussion

### 7.1. Statistical Analysis

In order to do a statistical analysis to identify the speakers' language preference, especially in closely related code-mixed datasets, we count the number of each sentiment tag in each language span; see Figure 4. The frequency of the Magahi span is high in the positive and neutral domains. Hindi is more prevalent in negative sentiment, whereas English is more frequent in mixed sentiment. While looking at the distribution of the tags across the language, we can deduce the following points:

1. Code-mixing affects the overall sentiment of the comments ( sentence level sentiment) as we can see that the percentage of positive and negative sentiment is higher than the neutral sentiment in language sentence level sentiment. In contrast, the distribution of neutral sentiment is high in language-specific span.

2. The speakers use more Magahi to express positive or neutral sentiments, which implies that the speakers are trying to leave a positive impression of the matrix language, which is Magahi in this case. It could also depend on the theme of the videos in which the comment is posted. For Example, a video on Magahi culture, folklore and movies will have a more positive labelled Magahi span to give listeners a positive impression of the Magahi language and culture.

3. The speakers use Hindi to express negative sentiments. This could be because the speakers emphasise their disagreement towards the non-Magahi community, especially in the videos, which discuss the domination of Hindi as a language and society over the Magahi community.

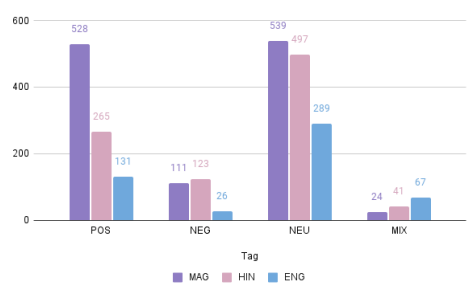

Figure 4: Distribution of the tags in the MHE dataset for language-specific span-level sentiment analysis.

### 7.2. Linguistic Analysis

The following section provides a linguistic analysis of the proposed dataset. We conducted the analysis manually to study the linguistic characteristics and understand the fundamental structure of code-mixing in closely related languages.

1. Like most closely related languages, Hindi and Magahi share a lexicon, possibly due to various reasons like phonetic similarity in the spelling across languages. For example, the letter 'u' represents *that* in Magahi and *You* in English. Similarly, many tokens are responsible for creating inconsistent tags. However, when the speakers try to express their strong emotions or gain attention, they code-mix a substantial number of functional words, including Wh-words, adjectives, pronouns, determiners, etc.

(1) E [language (lang) -"MAG"] Match me aapka dubbing matching ho gya Sirji [lang-"HIN"]

*Gloss*: this match in your-3SG.HON Dubbing match finish-PST AUX Sirji
*Translation*: Your dubbing got matched in this match Sirji

In Example 1, the determiner `E` is inserted in the utterance. This is when the speaker tries to emphasize a certain object, and in this example, that object is *match*.

(2)  बहुते [lang-"MAG"] सुंदर [lang-"HIN"] अपने मगही मे गीत गइली मन गदगद होगेल।
*Gloss*: very beautiful you-2SG.HON Magahi in song sing-PST.2SG.HON mind happy be-PST
*Translation*: You sang a very beautiful song in Magahi. It was mind-blowing.

The Example 2 shows the code-mixing between Magahi and Hindi where a Hindi adjective सुंदर (beautiful) is inserted in a Magahi sentence.

2. Insertion of the marker `-wa` and the numeral classifier `go` with the noun and act like a classifier language rather than a noun class like Hindi.

(3)  Ek [lang-"HIN/MAG"] du [lang-"MAG"] **go** [lang-"MAG"] aur [lang-"HIN"] likhiye [lang-"HIN"]
*Gloss*: one two go-CLF more write-PRS.2SG.HON
*Translation*: Write one or two more.

In Example 3 with the insertion of classifier `-go`, the text does not exhibit the number agreement in the noun morphology.

(4)  WOW haste haste [lang-"HIN"] mera [lang-"HIN"] pet**-wa** [lang-"MAG"]fat jayega [lang-"HIN"]
*Gloss*: WOW laugh-PRS.PROG.1SG my stomach-POSS hurt-PRS.PROG.1SG
*Translation*: My stomach hurts from laughing

In example 4, the noun `pet` is affixed with the familiarity marker `-wa`, presupposing that the noun is familiar to both the speaker and the listener.

3. When the speakers try to quote somebody or some famous expression, they code-mix as shown in Example 5. In the given example, the speaker recites the quote in Magahi, using the post-position `k`, thus making the quote in Magahi.

(5)  Once a legend said [lang-"ENG"] "Laura k sarkaar"[lang-"MAG"]
*Translation*: Once a legend said "evil government"

4. While talking about culture or regional traditions, the speakers often code-mix a lot as they are comfortable talking about their culture and tradition in their mother tongue.

(6)  Murna [label-"NAME"] ke [lang-"MAG"] khasi [label-"NAME"] khaiye [lang-"MAG"] ke [lang-"MAG"] hai [lang-"MAG"] Murna [label-"NAME"] kab [lang-"HIN"] hai [lang-"HIN"]☺
*Gloss*: Murna GEN mutton eat-PRS.1SG bePRS when bePRS
*Translation*: We have to get Murna's party, When is the Murna?

In Example 6, the speaker talks about a ritual called `Murna`. Traditionally, the person doing the rituals is supposed to throw a party, which is a very cultural aspect of Magahi society; therefore, the speaker code-mixes while discussing these cultural aspects. Moreover, this comment is also a perfect example of intra-sentential code-mixing.

5. When the speakers try to express their strong emotion of surprise or any other emotion, they are prone to code-mix between Magahi English as they insert interjections or exclamation in the text. For Example, the insertion of `WOW` at the beginning of the comment in example 4 express that the speaker is very happy and has a positive emotion.

(7)  Bhaiya ["HIN/MAG"] hum ["HIN"] kehrhe ["HIN"] hain ["HIN"] **ki** ["HIN/MAG"] E ["MAG"] sarkar ["NAME"] ke ["MAG"] sabhe ["MAG"] kala ["HIN/MAG"] cita ["HIN/MAG"] khol ["MAG"] da ["MAG"].
*Gloss*: Brother I say-1SG.PROG that-COMP this government evil-doing bring out
*Translation*: Brother, I am telling you to bring out all the evil-doing of this government.

Similarly, when the complementiser is used to express their strong emotion, the speakers tend to code-switch, as shown in example 7.

From this analysis, we tried pinpointing some of the linguistic properties of code-mixing in the dataset. However, these user-generated texts exhibit other characteristics like borrowing. The scenario where borrowing is taking place is very different from code-mixing. For example, the speakers borrow words from English in the expression

where they find it challenging to find the equivalent words in Hindi. For example:

(8) ई dislike कोन करलो हे साच हई हमनी अपन भासा काहे भुली
Transliteration: E dislike kaun karlo hae saach hai humni apen bhasa kahe bhuli
This dislike who do-PST AUX true we ours language why forget
Translation: Who disliked it? This is true; why will I forget my language?

In Example 8, the word `dislike` is being borrowed. It does not make any difference in the linguistic structure or the discourse if the word is replaced with an equivalent word. However, in YouTube text, words such as `like`, `subscribe`, `comment`, and `dislike` are easily borrowed from English, especially for Indian language speakers, as some of these words are complicated to type in the Devanagari scripts due to the complex typing format.

## 8. Conclusion

In this paper, we described the development process of the Magahi-Hindi-English code-mixed dataset annotated with sentiment. We analysed the speaker's language preference for a specific context with the help of a newly developed dataset. Moreover, we discussed some of the linguistic features that gave us insight into the linguistic structure of the code-mixed text. The analysis concluded that Magahi is used to express positive sentiments more than negative sentiments. These results do not agree with the previous studies (Agarwal et al., 2017; Rudra et al., 2016; Doğruöz et al., 2021), which state that the speakers prefer the first language to express negative sentiments or while swearing, which could be because of the code-mixing between two closely related languages i.e., Magahi and Hindi, and the fact that Hindi is dominant language in the Magahi spoken area affects the preference of language to shows different emotion.

We have also experimented with the baseline models using deep learning and SOTA transformer models. The results of the baseline models show that the multilingual model is relatively good in less-resourced and code-mixed settings. However, our future work will include model engineering to enhance the performance of SOTA so that we can capture the code-mixed representation well, which helps to capture the speakers' emotions efficiently.

## 9. Limitations

Due to the limited contents in Magahi, we could only collect a few texts. Some of the content is not publically accessible due to privacy reasons in the provided YouTube channels, which limits access to the data. To study language preference in closely related code-mixed scenarios, we need more studies to establish a final statement other than quantitative analysis.

## 10. Ethics Statement

All the data collected has been scrapped legally and validly, adhering to the provided guidelines. We developed the dataset under the ethical and legal framework of our university and have followed legal requirements as they apply to fair research use of such data. We have anonymised all comments and will remove any comments from the dataset on request should anything be found to be non-compliant. We followed proper data statement guidelines (Bender and Friedman, 2018) to annotate the dataset to establish the ethical issues. The data might contain strong language, which might be unsuitable for some applications. We will use the data only for research to preserve the users' privacy.